# Biogeography-Based Combinatorial Strategy for Efficient AUV Motion Planning and Task-Time Management

**Somaiyeh M.Zadeh, David MW Powers, Amirmehdi Yazdani**

School of Computer Science, Engineering and Mathematics,
Flinders University, Adelaide, Australia
somaiyeh.mahmoudzadeh@flinders.edu.au
david.powers@flinders.edu.au
amirmehdi.yazdani@flinders.edu.au

*Abstract* Autonomous Underwater Vehicles (AUVs) are capable of spending long periods of time for carrying out various underwater missions and marine tasks. In this paper, a novel conflict-free motion planning framework is introduced to enhance underwater vehicle's mission performance by completing maximum number of highest priority tasks in a limited time through a large scale waypoint cluttered operating field, and ensuring safe deployment during the mission. The proposed combinatorial route-path planner model takes the advantages of the biogeography-based optimization (BBO) algorithm toward satisfying objectives of both higher-lower level motion planners and guarantees maximization of the mission productivity for a single vehicle operation. The performance of the model is investigated under different scenarios including the particular cost constraints in time-varying operating fields. To show the reliability of the proposed model, performance of each motion planner assessed separately and then statistical analysis is undertaken to evaluate the total performance of the entire model. The simulation results indicate the stability of the contributed model and its feasible application for real experiments.

## 1 Introduction

Recent advancements in sensor technology and embedded computer systems has opened a new dimension in underwater autonomy and made autonomous underwater vehicles (AUVs) more conspicuous for handling long-range underwater missions. However, today's AUVs' endurance still remains restricted to short ranges due to battery restrictions. Autonomous deployment of AUV in a vast, unfamiliar and dynamic underwater environment is a complicated process, specifically when the AUV is obligated to perform prompt reaction to environmental changes, where mostly real-time information is not known. In this context, accurate motion planning strategy promotes the vehicles autonomy in terms of carrying out different tasks in a pre-specified time interval while ensure the vehicles' safety at all-time spans. This highly tied with the vehicles' autonomy in high-low level motion planning and its ability in prioritizing the available tasks to be completed in a limited time that battery capacity allows, while it is concurrently guided toward its final destination in a graph-like terrain; which is analogous to both knapsack and travelling salesman problems (TSP) at the same time. Hereupon, the vehicle has to effectively use the available time for a series of deployment in a long mission, which tightly depends on the optimality of the selected route between start and destination in a waypoint cluttered terrain. Accurate motion planning strategy provides a careful use of the AUV for hitting the mission goals while avoiding the common pitfalls associated with unknown environments.

Many attempts have been carried out in scope of single or multiple vehicles motion planning and task allocation using different strategies. Some instances of route planning systems applications are in the areas of traffic control [1], real time routing and trip planning [2], modelling the transportation network to find the shortest paths [3], etc. The application of genetic algorithm is investigated by [4] in a dynamic route guidance system. A behavior-based controller joint with waypoint tracking scheme presented by [5] for an AUV guidance in large scale underwater environment. An artificial potential field approach employed by [6] for energy efficient routing of the underwater vehicle. An energy efficient fuzzy based approach using priori known wind information in a graph-like terrain is presented by [7] for unmanned aerial vehicle route planning. A large scale AUV routing and task assignment joint problem has been investigated in [8] by transforming the problem space into a NP-hard graph context, in which the heuristic search nature of GA and PSO employed to find the best series of waypoints. Later on, a task assign route planning model based on BBO and PSO algorithms is introduced by [9] for a time efficient routing of the AUV in a semi-dynamic operation network, where location of some the waypoints are changed by time in a bounded area.

Remarkable efforts also have been devoted on unmanned vehicles optimum path planning in recent years. Normally, the energy cost agrees the path time. A real-time collision-free robot-path planning based on a dynamic-programming shortest path algorithm is proposed by [10]. A higher geometry maze routing algorithm is investigated by [11] for optimal path planning and navigating a mobile rectangular robot among obstacles. An A* algorithm is applied by [12] for AUV path planning problem taking variable vehicle speeds into account. In another research [13] also, an A*-based path planer employed to find path with minimum energy consumption considering ocean variations. The A* acts more efficiently because of its heuristic searching capability; however, it suffers from expensive computational cost in larger search spaces. A mixed integer linear programming (MILP) is implemented for handling multiple AUVs path planning problem by [14]. A non-linear least squares optimization technique is employed by [15] for AUV path planning through the Hudson River. Path planning based on deterministic methods is carried out repeating a set of predefined steps that search for the best fitted solution to the objectives [16]. The deterministic methods are inaccurate in large sized problems as its computational time increases exponentially with the problem size. The meta-heuristic approach is a good alternative with fast computational speed, specifically in dealing with NP-hard multi-objective optimization problems [17,18].

In most of previous researches, vehicles routing strategies mostly have been investigated for vehicles' mission planning, task allocation and time scheduling purposes [1-9], while the path planning strategies have been proposed to find the optimum safe path to the predefined target point [10-16]. Combining these two strategies helps to cover shortcoming associated with each of them and furnishes objectives of both vehicle task assign/routing and collision free path planning problem taking uncertainty and dynamicity of the environment into account.

For having a reliable underwater missions in a large scale environment in presence of considerable amount of uncertainty, this paper contributes a combinatorial framework containing an efficient graph rout planning accompanying with real-time path planning that guarantees a successful operation of the single vehicle. The graph route planner, in this context, is capable of generating the best fitted route to the available time that includes the maximum number of highest priority tasks and ensure the AUV reaches to the destination on-time; hence, as an higher level motion planner it is in charge of optimal distribution of tasks that sparsely assigned to edges of an operation network to guide the vehicle from the given starting point to the target of interest. Beside an efficient route planner, the path generator is employed to find a conflict-free trajectory in a smaller scale that is designed to be fast enough to handle sudden changes of the environment and safely guide the vehicle through the specified waypoints with minimum time/energy cost. A constant interaction is flowing between the path and route planners by feed backing the situational awareness of the surrounding operating filed form the local path planner to the graph route planner for making a decision on requisitions of re-planning process. Hence, re-planning is performed to generate a new optimum route toward the destination according to the last update of decision variables. This process continues until the AUV reaches to the end point. As the best of authors' knowledge, the contributed combinatorial strategy is a novel prospective to ordinary motion planning systems that advances the system to cover broader requirement of an autonomous operations by trade-off within the problem constraints, managing the time, having efficient mission by proper prioritizing the tasks, and ensuring safe deployment during the mission. The proposed strategy is also efficient in computational time that enhances the system's real-time performance.

In the core of the proposed strategy, both path and graph route planners take the advantages of Biogeography-based Optimization (BBO) algorithm toward satisfying addressed objectives in this research. A similar hierarchal model of route-path planning with slight difference in problem formulation is also implemented applying PSO algorithm for both higher-lower level planners [19]. Obviously acquiring the optimal solutions for non-deterministic polynomial-time hard problems is a computationally arduous issue and currently there is no polynomial time algorithm that can solve an NP-hard problem of even moderate size. Besides, obtaining the accurate optimal solution is only applicable for completely known environments without any uncertainty, while the modelled environment by this paper corresponds to a highly uncertain dynamic environment. Considering synthetic characteristic of AUV's task allocation/routing problem, which generalizes both TSP and Knapsack problems, and also taking NP-hard complexity of the path planning problem in to account, the BBO algorithm is applied by this research as one of the fastest meta–heuristic algorithms introduced for solving NP-hard problems [20]. The argument for using the BBO in solving NP-hard problems is strong enough due to its well adaption to multidimensional spaces and proper scaling with multi-objective problems, which in the proceeding research has been shown to produce near optimum solutions with high probability. A particular characteristic of the BBO algorithm is that the solutions of one generation get transferred to the next and the primary population never get discarded but get modified by migration, which this issue promotes the exploitation ability of the algorithm. This algorithm also employs mutation operator to promote the diversity of the population that propel the individuals toward global optima.

In order to examine the capability and efficiency of the proposed framework, multiple simulations have been carried out and implemented in MATLAB®2014 considering different situations of the dynamic uncertain environments. The organization of the paper is as follows. An overview of the Biogeography-Based Optimization Algorithm is provided in section 2. The formulation and implementation of the BBO-based graph route planner is demonstrated in section 3. In section 4, the BBO-based path planer is formulated. Evaluation criterion of the entire combinatorial model is discussed in section 5. The discussion on simulation results are provided in section 6. And, the section 7 concludes the paper.

## 2 Overview of Biogeography-based Optimization

The BBO is an evolutionary optimization technique developed based on the equilibrium theory of island biogeographical concept [21]. The basic idea of the algorithm inspired by the immigration, emigration, and rate of change in the number of species in an island. The geographically isolated islands are known as habitats that correspond to problem solutions, which in the proceeding research each solution corresponds to a candidate path/route generated by the path and route planners. Each candidate solution in BBO has a quantitative performance index representing the fitness of the solution called habitat suitability index (HSI). High HSI solutions refer to islands with more suitable habitation. Habitability is related to some qualitative factors known as Suitability Index Variables (SIVs), which is a vector of integers initialized randomly in advance. Therefore, each particular solution (habitat) of $h_i$ has a design variable of SIV, emigration rate of μ and immigration rate of λ. The emigration and immigration rates directly affect the population size and tend to improve the solutions. Each solution of the population should be evaluated before starting the optimization process. A poor solution has higher immigration rate of λ and lower emigration rate of μ. The immigration rate λ is used to probabilistically modify the SIV of a selected solution $h_i$. Then emigration rates μ of the other solutions are considered and one of them probabilistically selected to migrate its SIV to solution $h_i$. This process known as migration in BBO. Afterward, the mutation operation is carried out that tends to increase the diversity of the population and propels the individuals toward the global optima. Each given solution $h_i$ is modified according to probability of $P_s(t)$ that is the probability of existence of the $S$ species at time $t$ in habitat $h_i$. To have $S$ species at time $(t+\Delta t)$ in a specific habitat $h_i$, one of the following conditions must hold:

[1] The $S$ species exist in $h_i$ at $t$, and no emigration or immigration is occurred from $t$ to $(t+\Delta t)$;
[2] One species immigrated onto an island already occupied by ($S-1$) species at $t$;
[3] One species emigrated from an island occupied by ($S+1$) species at $t$;

In mathematical representation of BBO, the probability $P_s(t+\Delta t)$ gives the change in number of species after time $\Delta t$, calculated by

$$\forall h_i(t) \ \exists \lambda_S, \mu_S, P_S(t)$$
$$\begin{cases} \lambda_S = I - \dfrac{I \times S}{S_{\max}} \\ \mu_S = \dfrac{E \times S}{S_{\max}} \end{cases} \xrightarrow{if\ E=I} \lambda_S + \mu_S = E \qquad (1)$$
$$P_S(t+\Delta t) = P_S(t)(1 - \lambda_S \Delta t - \mu_S \Delta t) + P_{S-1} \lambda_{S-1} \Delta t + P_{S+1} \mu_{S+1} \Delta t$$

where $\lambda_s$ and $\mu_s$ are the immigration and emigration rates when there are S species in the habitat $h_i$. $I$ and $E$ are the maximum immigration and emigration rates set by the user. Maximum emigration rate ($E$) occurs if all species $S_{max}$ are collected in a habitat. As habitat suitability improves, the number of its species and emigration increases, and the immigration rate decreases. The probability of more than one emigration/immigration can be neglected by assuming a very small $\Delta t$. If time $\Delta t$ is negligible, as $\Delta t \to 0$, $P_s$ is calculated by

$$\dot{P}_S = \begin{cases} -(\lambda_S + \mu_S)P_S + P_{S+1}\lambda_{S+1} & S = 0 \\ -(\lambda_S + \mu_S)P_S + P_{S+1}\mu_{S+1} + P_{S-1}\lambda_{S-1} & 1 \le S \le S_{\max} - 1 \\ -(\lambda_S + \mu_S)P_S + P_{S-1}\mu_{S-1} & S = S_{\max} \end{cases} \qquad (2)$$

Mutation is required for solution with low probability, while solution with high probability is less likely to mutate. Hence, the mutation rate $m(S)$ is inversely proportional to probability of the solution $P_s$.

$$m(S) = m_{\max}\left[\dfrac{1 - P_S}{P_{\max}}\right] \qquad (3)$$

where $m_{max}$ is the maximum mutation rate defined by user, $P_{max}$ is probability of the habitat with maximum number of species $S_{max}$. The BBO is applied on both path planning and route planning problems. The BBO is well suited for solving vehicles path planning problem due to continuous nature of this problem [18,20].

### 3 Formalization of BBO Global Route Planning

For a single vehicle's operation, it is not possible to cover all tasks in a single mission in a large scale operation area. Therefore, available tasks are prioritized in a way that selected edges (tasks) of the graph can govern the AUV to the destination; this is analogous to a joint discrete and syndetic space problem that should be considered simultaneously. In this context, the proposed route planning problem can be modelled as a multi-objective optimization problem. The BBO is a particular type of stochastic search algorithm representing problem solving technique based on Biogeographical evolution and scales well with complex and multi-objective problems. Exploiting a priori knowledge of the underwater environment, the initial step is to transform the problem space into a graph problem as depicted in *Fig.*6. The vehicle starts its mission from initial position of $WP^1:(x_1,y_1,z_1)$ and should pass sufficient number of waypoints to reach the destination $WP^D:(x_D,y_D,z_D)$. The waypoints' location are randomized according to a uniform distribution of $\sim U(0,10000)$ for $WP^i_{x,y}$ and $\sim U(0,100)$ for $WP^i_z$. Waypoints in the terrain are connected with edge $q_i$ from a set of $q=\{q_1,...,q_m\}$, where $m$ is the number of edges in the graph. Each edge $q_i$ from graph is assigned with a specific task from a set of Task=$\{Task_1,...,Task_k\}$ $k \in m$. Each task involves parameters of priority $\rho$, and absolute completion time $\delta$ that are initialized randomly in advance. If the route is $R_i=(x_1,y_1,z_1,...,x_i,y_i,z_i,...,x_D,y_D,z_D)$, where $WP^i:(x_i,y_i,z_i)$ is the coordinate of any arbitrary waypoint in geographical frame, the route travelled time is calculated as follows.

$$q_{ij}: \begin{cases} d_{ij} = \sqrt{(x_j - x_i)^2 + (y_j - y_i)^2 + (z_j - z_i)^2} \\ t_{ij} = \dfrac{d_{ij}}{V_{AUV}} + \delta_{T_{ij}} \end{cases} \ ; \ Task_{q_{ij}}: \begin{cases} \rho_{T_{ij}} \\ \delta_{T_{ij}} \end{cases} \qquad (4)$$

where $t_{ij}$ is the required time to pass the distance $d_{ij}$ between two waypoints of $WP^i$ and $WP^j$ that includes the corresponding task's completion time $\delta_{Tij}$ and $\rho_{Tij}$ is the task priority.

The second step is to generate feasible primary routes as initial population of habitats for BBO optimization process. Developing a suitable coding scheme for habitats representation is the most critical step in implementing BBO framework. Hence, efficient representation of the routes and encoding them correctly into the habitats has direct impact on overall performance of the algorithm and optimality of the solutions. Habitats in the proposed BBO should correspond to a feasible route including a sequence of nodes. Feasibility of a generated route gets assessed by following criteria:

- A valid route should be commenced and ended with predefined start and target nodes.
- The generated route should not include edges that are not presented in the graph.
- The multiple appearance of the same node in a route makes it invalid because this issue implies wasting time repeating a task.
- The route should not traverse an edge for more than once.
- The route travel time should not exceed the maximum range of AUVs' total available time.

This research conducts a priority based strategy to generate feasible routes. To this purpose, a randomly initialized priority vector is assigned to sequence of nodes. Adjacency information of the operation network and generated priority vector get conducted for proper node selection of a feasible route. To prevent generating infeasible routes some modifications have been applied. To generate a feasible route in a graph based on topological information, each node take positive or negative priority values in the specified range of [-100,100]. Adjacency relations are used for adding nodes to a specific route, so that nodes are added to the route sequence one by one according to priority vector and adjacency matrix. The first node is selected and added to the sequence as the start position. Then from adjacency matrix the connected nodes to node-1 is considered. The node with the highest priority in this sequence is selected and added to the route sequence as the next visited node. The selected node in a route sequence gets a large negative priority value that prevents repeated visits to a node. Then, the visited edges get eliminated from the Adjacency matrix. So that, the selected edge will not be a candidate for future selection. This issue reduces the memory usage and time complexity for large and complex graphs. This procedure continue until a legitimate route is built (destination visited). To satisfy the termination criteria of a feasible route, if the route ends with a non-destination node and/or the length of the route exceeds the number of existed nodes in the graph, the last node of the sequence will be replaced by index of the destination node. The process of the BBO based global route planner is summarized in following flowchart in *Fig.1*.

After a bunch of feasible routes are generated and habitats population initialized, the optimization process get started to find the optimum global route through the given waypoints according to the flowchart in *Fig.*1. The goal is to find the route that covers maximum number of highest priority tasks in a time interval that battery's capacity allows. The problem involves multiple objectives that should be satisfied during the optimization process. In this regard, the objective function of the route planner is defined in a form of hybrid cost function comprising weighted functions that are required to be maximized or minimized. The total route cost function is formulated in section 5. In this model, the route travel time should be approached to the mission available time (means maximizing the use of available time).

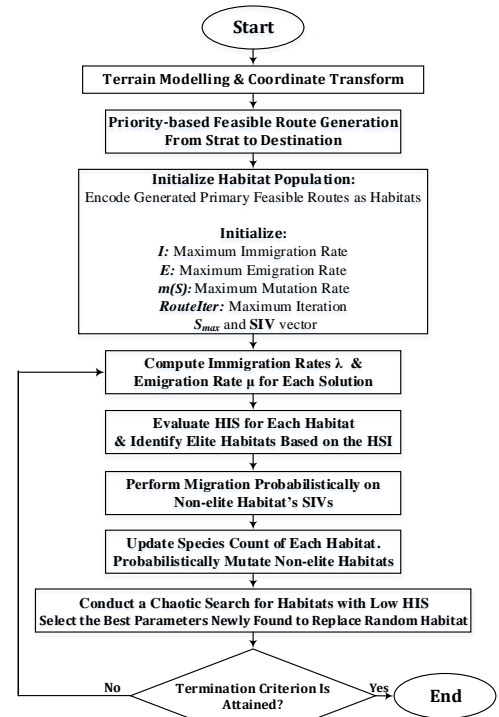

Fig.1. Process of the BBO based global route planning

$$T_{Route} = \sum_{\substack{i=0 \\ j \neq i}}^{n} l q_{ij} t_{ij} = \sum_{\substack{i=0 \\ j \neq i}}^{n} l q_{ij} \left( \frac{d_{ij}}{V_{AUV}} + \delta_{Tij} \right), \quad l \in \{0,1\} \quad (5)$$

$$\min(|T_{Route} - T_{Available}|)$$
$$s.t. \quad (6)$$
$$\max(T_{Route}) < T_{Available}$$

where $T_{Route}$ is the required time to pass the route, $T_{Available}$ is the total mission time, $l$ is the selection variable.

## 4 Formalization of BBO Local Path Planning

The local path planner operates in context of the global route planner and concurrently tends to generate a safe collision-free path between pairs of waypoints encountering dynamicity of the environment. In this section, conceptual and mathematical representation of the local path planning framework is presented. The path planning is an optimization problem in which the main objective is to find a time optimum collision-free local path $\wp_i$ (shortest path) between specific pair of waypoints in the presence different types of uncertain obstacles. The resultant path should be safe and flyable (feasible). Dynamicity of the operation environment $\Gamma_{3D}$ addressed encountering different type of static and floating obstacles $\Theta$ with uncertain position and velocity, where the floating obstacles affected by current flow. The proposed path planner in this study, generates the potential trajectories $\wp_i:\{\wp_1,\wp_2,...\}$ using B-Spline curves captured from a set of control points like $\vartheta = \{\vartheta_1, \vartheta_2, ..., \vartheta_i, ..., \vartheta_n\}$ in the problem space with coordinates of $\vartheta_1:(x_1,y_1,z_1),..., \vartheta_n:(x_n,y_n,z_n)$, where $n$ is the number of corresponding control points. These control points play a substantial role in determining the optimal path. The mathematical description of the B-Spline coordinates is given by (7):

$$\left. \begin{array}{l} X(t)=\sum_{i=1}^{n}\vartheta_{x(i)}B_{i,K}(t) \\ Y(t)=\sum_{i=1}^{n}\vartheta_{y(i)}B_{i,K}(t) \\ Z(t)=\sum_{i=1}^{n}\vartheta_{z(i)}B_{i,K}(t) \end{array} \right\} \mapsto \forall \wp, \quad \wp \approx \sum_{1}^{|\wp|}\vartheta_{i+1}-\vartheta_i, \quad (7)$$

$$\wp^i_{x,y,z} = \sum_{x_s,y_s,z_s}^{|\wp|}\sqrt{(\vartheta_{x(i+1)}-\vartheta_{x(i)})^2+(\vartheta_{y(i+1)}-\vartheta_{y(i)})^2+(\vartheta_{z(i+1)}-\vartheta_{z(i)})^2}$$

here, the $X(t), Y(t)$, and $Z(t)$ give the vehicles position along the path at time $t$, the $B_{i,K}(t)$ is the curve blending functions, $K$ is the order of the curve represents the smoothness of the curve, where bigger $K$ correspond to smoother curves. For further information refer to [22]. All control points should be located in respective search region constrained to predefined bounds of $\beta^i_\vartheta=[U^i_\vartheta,L^i_\vartheta]$. If $\vartheta_i:[x(i), y(i),z(i)]$ represent one control point in Cartesian coordinates, he lower bound $L^i_\vartheta$ t; and the upper bound $U^i_\vartheta$ of all control points at $(x$-$y$-$z)$ coordinates is calculated by (8,9). Afterward, each control point $\vartheta_i$ is generated from (10):

$$\begin{array}{l} L_{\vartheta(x)}=[\vartheta_{x(0)},\vartheta_{x(1)},...,\vartheta_{x(i-1)},...,\vartheta_{x(n-1)}], \\ L_{\vartheta(y)}=[\vartheta_{y(0)},\vartheta_{y(1)},...,\vartheta_{y(i-1)},...,\vartheta_{y(n-1)}], \\ L_{\vartheta(z)}=[\vartheta_{z(0)},\vartheta_{z(1)},...,\vartheta_{z(i-1)},...,\vartheta_{z(n-1)}], \end{array} \quad (8)$$

$$\begin{array}{l} U_{\vartheta(x)}=[\vartheta_{x(1)},\vartheta_{x(2)},...,\vartheta_{x(i)},...,\vartheta_{x(n)}], \\ U_{\vartheta(y)}=[\vartheta_{y(1)},\vartheta_{y(2)},...,\vartheta_{y(i)},...,\vartheta_{y(n)}], \\ U_{\vartheta(z)}=[\vartheta_{z(1)},\vartheta_{z(2)},...,\vartheta_{z(i)},...,\vartheta_{z(n)}], \end{array} \quad (9)$$

$$\begin{array}{l} \vartheta_{x(i)}=L^i_{\vartheta(x)}+Rand^x_i(U^i_{\vartheta(x)}-L^i_{\vartheta(x)}) \\ \vartheta_{y(i)}=L^i_{\vartheta(y)}+Rand^y_i(U^i_{\vartheta(y)}-L^i_{\vartheta(y)}) \\ \vartheta_{z(i)}=L^i_{\vartheta(z)}+Rand^z_i(U^i_{\vartheta(z)}-L^i_{\vartheta(z)}) \end{array} \quad (10)$$

In the proposed BBO based path planning problem, each habitat $h_i$ corresponds to the coordinates of the B-spline control points $\{\vartheta_1,...,\vartheta_n\}$ that is utilised in path generation. Each habitat $h_i$ has a HSI index and represented by real vector of $n$-dimension that initialized randomly in advance. A habitat is a vector of $n$ SIVs ($h_i:\{\chi_1, \chi_2, ..., \chi_n\}$) generated randomly and defined as a parameter to be optimized.

Afterward the migration and mutation operations are carried out to lead the habitats toward the optimal solution. As the BBO algorithm iterates, each habitat gets attracted toward its respective best position based on habitat suitability index variable (SIV). The pseudo code of the BBO algorithm and its mechanism on path planning process is provided in *Fig.2*.

The path planner is applied in a small scale area, and the AUV is considered to have constant thrust power; therefore, the battery usage for a path is a constant function of the time and distance travelled. Performance of the generated path is evaluated based on overall collision avoidance capability and time consumption which is proportional to energy consumption and travelled distance. *Fig.*3 illustrates the schematic of the AUV path planning process.

```
Steps in BBO Algorithm for Path Planning
Initialize a set of solutions as initial habitat population
  ▪ Assign each habitat h_i:{ χ_1, ..., χ_n} with a candidate path of ℘_i:{ϑ_1,...,ϑ_n}
  ▪ Choose appropriate parameters for the population size nPop
  ▪ Set the number of control-points (n) that used to generate the B-Spline path
  ▪ Set the maximum number of iteration Iter_max
  ▪ Assign maximum immigration and emigration rate (I, E)
  ▪ Assign maximum mutation rate m(S)
  ▪ Set S_max and SIV vector
For t=1 to Iter_max
  Compute immigration rates λ and emigration rate μ for each solution
      λ_S(t) = I*(1 - S/S_max);  μ_S(t) = E*(S/S_max)
  Evaluate the fitness (HSI) of each habitat and identify Elite Habitats based on HIS
  Modify habitats based on λ and μ (Migration):
      P_S(t+Δt) = P_S(t)(1 - λ_S Δt - μ_S Δt) + P_{S-1} λ_{S-1} Δt + P_{S+1} μ_{S+1} Δt
  For i=1 to nPop
    Use λ_i to probabilistically decide whether immigrate to habitat h_i
    if rand(0,1) < λ_i
      For j=1 to nPop
        Select the emigrating habitat h_j with probability ∝ μ_j
        if rand(0,1) < μ_j
          Replace a randomly selected SIV variable of h_i with its corresponding value in h_j
        end (if)
      end (For)
    end (if)
  end (For)
  Carry out the mutation based on probability by m(S) = m_max [ (1-P_S)/P_max ]
  Transfer the best solution in the population from one generation to the next
end (For)
Output the best habitat and its correlated path as the optimal solution
```

Fig .2. BBO path planning pseudo code

In terms of collision avoidance, obstacle's velocity vectors and coordinates can be measured by the sonar sensors with a certain uncertainty modelled with a Gaussian distributions. The state of obstacle(s) continuously measured and sent to state estimator to provide the estimation of the future states of the obstacles for the local path planner. The state predictor estimates the obstacles behaviour during the vehicles deployment in specified operation window. Four different type of obstacles are conducted in this study to evaluate the performance of the proposed path planner, in which the obstacle is presented by three components of position, radius and uncertainty $\Theta_{(i)}:(\Theta_p,\Theta_r,\Theta_{Ur})$. The first kind is static known obstacle that its location is known and can be obtained from the offline map. No uncertainty growth is considered for position of this type. The second type, is also classified as the Quasi-static obstacles that usually known as no flying zones. The obstacles in this category has an uncertain radius varied in a specified bound with a distribution of $\sim \mathcal{N}(\Theta_p,\sigma_0)$, where the value of $\Theta_r$ in each iteration is independent of its previous value. Self-motivated moving obstacle, is the third type that has a motivated velocity that shift it from a position A to position B. Therefore, its position changes to a random direction with an uncertainty rate proportional to time, given in (11). The last type is the moving obstacle that affected by current force and also have a self-motivated velocity to a random direction, denoted by (11,12). Here, the effect of current presented by uncertainty propagation proportional to current magnitude $U_R^C=|V_C|\sim (0,0.3)$ that radiating out from the center of the obstacle in a circular format presented by *Fig.*4.

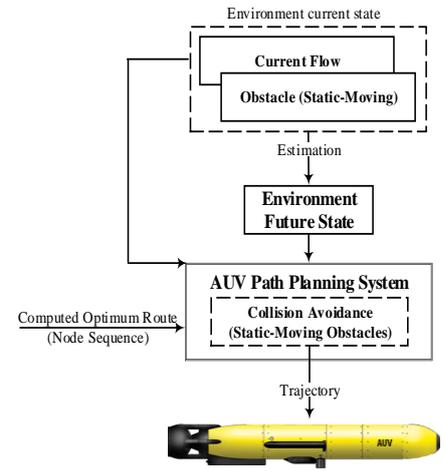

**Fig.3.** The operation diagram of the AUV path planning

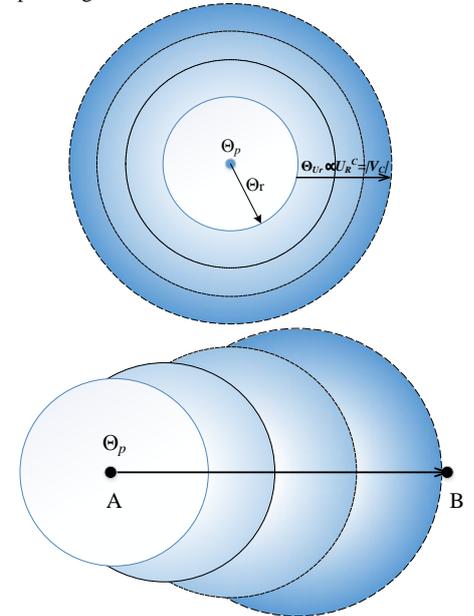

**Fig.4.** The graphical presentation of uncertain floating/moving obstacles

$$\Theta_p(t)=\Theta_p(t-1)\pm \mathbf{U}(\Theta_{p_0},\sigma) \tag{11}$$

$$\Theta_r(t)=B_1\Theta_r(t-1)+B_2X_{(i-1)}+B_3\Theta_{Ur}$$
$$B_1=\begin{bmatrix}1 & U_R^C(t) & 0\\ 0 & 1 & 0\\ 0 & 0 & 1\end{bmatrix}, B_2=\begin{bmatrix}0\\1\\1\end{bmatrix}, B_3=\begin{bmatrix}0\\0\\U_R^C(t)\end{bmatrix} \tag{12}$$

where $\Theta_{Ur}\sim\sigma$ is the rate of change in objects position, and $X_{(t-1)}\sim\mathcal{N}(\Theta_p,\sigma_0)$ is the Gaussian normal distribution that assigned to each obstacle and gets updated at each iteration *t*. In all cases, obstacle's position $\Theta_p$ initialized using normal distribution of $\sim (0,\sigma^2)$ bounded to position of start and target waypoints $WP^a_{x,y,z}<\Theta_p<WP^b_{x,y,z}$. Therefore the obstacles position $\Theta_p$ has a truncated normal distribution, where its probability density function defined as follows:

$$\Theta^i_p\in(WP^a_{x,y,z},WP^b_{x,y,z})-\Theta^i_r$$
$$f(\Theta^i_p;0,\Theta^i_r,WP^a_{x,y,z},WP^b_{x,y,z})= \tag{13}$$
$$\frac{\Theta^i_p}{(\Theta^i_r)^2}\bigg/\left(\frac{WP^b_{x,y,z}-\Theta^i_p}{\Theta^i_r}\right)-\left(\frac{WP^a_{x,y,z}-\Theta^i_p}{\Theta^i_r}\right)$$

The obstacle radius initialized using a Gaussian normal distribution of $\sim (0,100)$. This operating zone shifts to next pair of waypoints in sequence provided by the graph route planner. To evaluate the path $\wp_i$, the path cost function defined based on required time to travel along the path segments between two waypoints ($T_{path\text{-}flight}$).

$$\forall \wp,\quad \wp\approx \sum_1^{|\wp|}\vartheta_{i+1}-\vartheta_i,$$
$$\wp^i_{x,y,z}=\sum_{x_s,y_s,z_s}^{|\wp|}\sqrt{(\vartheta_{x(i+1)}-\vartheta_{x(i)})^2+(\vartheta_{y(i+1)}-\vartheta_{y(i)})^2+(\vartheta_{z(i+1)}-\vartheta_{z(i)})^2} \tag{14}$$
$$L_\wp=\sum_{x_s,y_s,z_s}^{|\wp|}\sqrt{(\vartheta_{x(i+1)}-\vartheta_{x(i)})^2+(\vartheta_{y(i+1)}-\vartheta_{y(i)})^2+(\vartheta_{z(i+1)}-\vartheta_{z(i)})^2}$$
$$Cost_\wp=L_\wp$$

$$T_{path\text{-}flight}=\sum_1^n t_i=\sum_1^{|\wp|}\frac{|\vartheta_{i+1}(t)-\vartheta_i(t)|}{|V_{AUV}|} \tag{15}$$

$$Cost_\wp=T_{path\text{-}flight}$$
$$s.t.$$
$$\forall j\in\{0,...,|\wp|\}\Rightarrow \vartheta_j^{t_j}\notin \Theta(t_j)\cup \Gamma_{3D} \tag{16}$$
$$\text{and } j\notin \bigcup_{N\Theta}\Theta(\Theta_p,\Theta_r,\Theta_{Ur})$$

where $j$ is any arbitrary point from the generated path, $Cost_\wp$ is the path cost function. The corresponding generated path shouldn't cross the forbidden area covered by obstacle $\Theta_{(N\Theta)}:(\Theta_p,\Theta_r,\Theta_{Ur})$, where $N_\Theta$ is number of obstacles. The generated path gets penalty value for colliding any obstacle. This process in general promotes algorithms evolution capability toward generating the feasible solution.

## 5 Evaluation of Combined BBO Motion Planning Model

As mentioned earlier, the local path planner operates in context of the global route planner and concurrently tends to generate the safe collision-free path between pairs of waypoints encountering dynamicity of the environment. After visiting each waypoint, the path absolute time $T_{path\text{-}flight}$ is calculated at the end of the path. This generated path time $T_{path\text{-}fligh}$, then is compared to the expected time $T_{Expected}$ for passing the distance between specified pair of waypoints. If $T_{path\text{-}flight}$ gets smaller value than $T_{Expected}$, it means no unexpected difficulty occurred during this course and vehicle can continue its travel along the provided route. However, if $T_{path\text{-}flight}$ exceeds the $T_{Expected}$, it means AUV has faced a challenge during its deployment. Obviously, a certain amount of the mission available time $T_{Avaliable}$ is wasted for copping the mentioned difficulties and therefore, the previously defined optimum route cannot be optimum anymore. In such a situation it is essential to re-plan a new optimum route according mission updates. Hence, in the third problem it is focused on route re-planning based on environmental changes and time updates during vehicles deployment. The process of the re-planning is clarified in *Fig*.5.

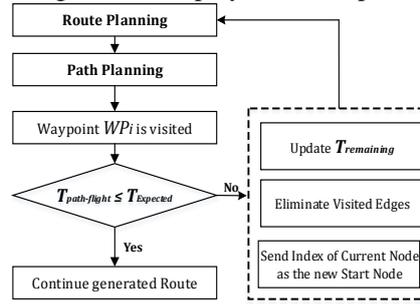

**Fig.5.** Requestion for replanning process

Another issue that has been taken into consideration in this situation is the computational burden of re-planning process. If after occurring any unexpected event the previously found optimum route is ignored and the global route planner is recalled to find the new optimum route from the predefined start point to the destination, the re-planning scheme does not guarantee at least a quasi-real-time solution as it embodies considerable computational burden. A beneficial approach for significant reducing the computational burden of re-planning is that if re-routing is required at any situation based on the updating of $T_{Avaliable}$, the passed edges get eliminated from the operation network (so the search space shrinks); and the location of the present waypoint is considered as a new start position for both local and global motion planners. Afterward, the global route planner tends to find the optimum route based on new information and updated network topology. For instance in *Fig*.6, the initial optimum route is a sequence of waypoints {1-2-5-7-12-15-13-16-11-17-D} and after re-planning, it is replaced by a new sequence of {15-18-16-D}. During deployment between two waypoints, the local path planner can incorporate dynamic changes of the environment. This process continues until mission ends and vehicle reaches the required destination point. The trade-off between available mission time and mission objectives is critical issue that can be adaptability carried out by route planner. Hence, it should be fast enough to track environmental changes and promptly re-plan a new route fitted to updated available time. A schematic representation of the combinatorial strategy is given by *Fig*.7.

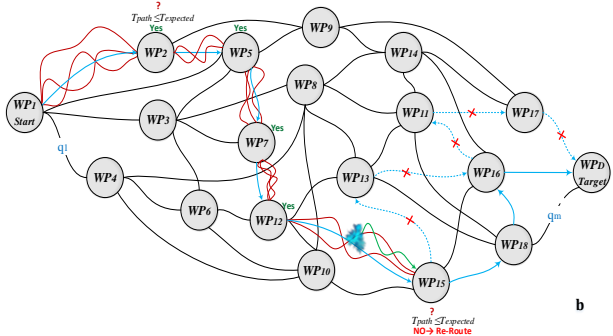

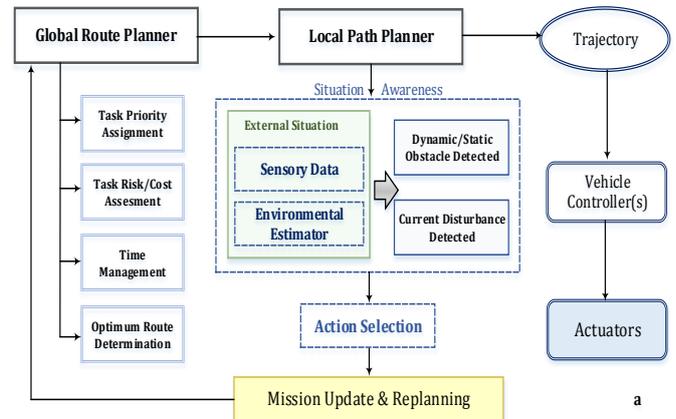

**Fig.6.** A graph representation of operating area covered by waypoints and local/global motion planning and re-planning process.

**Fig.7.** Process of the proposed combinatorial strategy for dynamic guidance of AUV.

The route cost has direct relation to the passing distance among each pair of selected waypoints with respect to equations (4-6). Hence, the path cost $Cost_\wp$ for any optimum local path gets used in the context of the graph route planner. The model is seeking an optimal solution in the sense of the best combination of path, route, and task cost. After visiting each waypoint, the re-planning criteria is investigated. A computation cost encountered any time that re-planning is required. Thus, total cost for the model defined as:

$$Cost_{Route} = \left| \sum_{\substack{i=0 \\ j \neq i}}^{n} lq_{ij}\left(Cost_{\wp ij} + \delta_{T_{ij}}\right) - T_{Available} \right| + \sum_{\substack{i=0 \\ j \neq i}}^{n} lq_{ij}/\rho_{T_{ij}} + \sum_{1}^{r} T_{compute} \quad , \quad l \in \{0,1\}$$

s.t.
$\forall Route_i \Rightarrow \max(T_{Route}) < T_{Available}$  (17)

Where, $\delta_{Tij}$ and $\rho_{Tij}$ represent completion time and priority value of the task assigned to $q_{ij}$. $Cost_{Route}$ is a function of path cost $Cost_\wp$ encountering task complettion time and computational time $T_{compute}$ required for checking the replanning criteria, and summation of priority of tasks included in the generated route. $r$ is the number of repeating the replanning procedure.

## 6 Discussion of Simulation Results

In this section, the simulation results regarding the performance evaluation of the proposed framework are demonstrated. The main purpose, is to analyze the performance of each motion planner and functionality of the entire framework, in terms of increasing mission productivity (task assignment and time management) along with ensuring vehicles safety during the mission. To verify the efficiency of the proposed strategy, first the efficiency of the local path planner is individually assessed; afterward, the performance of the global route planner is investigated; and finally in the last section, overall coherence of the framework in terms of mission timing and accurate cooperation of global and local planners is investigated and evaluated. For the purposes of this study, the optimization problem has been performed on a desktop PC with an Intel i7 3.40 GHz quad-core processor in MATLAB® R2014a.

### A. *Evaluation of the Path Planner*

It is assumed the AUV is traveling with constant velocity $V_{AUV}$ between two waypoints. The ocean environment has been modelled as a three dimensional environment $\Gamma_{3D}$ comprising known static, uncertain static, and floating/moving obstacles. In path planning simulation, obstacles are generated randomly from different categories and configured individually based on given relations in section 2. These assumptions plays an important role in efficient path planning and copping with environmental dynamic changes. The path planner in this study, generates the potential AUV trajectories using B-Spline curves captured from a set of control points. The fitness of the generated path is evaluated by equation (16). Three different scenarios are implemented in the simulation environment to assess the accuracy of the proposed local path planner. In the first one, the AUV starts its deployment in a pure static operating filed containing static known and static uncertain obstacles, where the vehicle is required to pass the shortest collision free distance to reach to the specified target waypoint. Making the AUV's mission more challenging, in the second scenario, the robustness of the method is tested in a dynamic environment including moving obstacles with self-motivated velocity moving in a random direction. In the third scenario, the mission becomes more complicated by adding the impact of the current force in a context of floating obstacles with uncertain position. The purpose of increasing complexity of the obstacles is to evaluate sustainability of the path planner's performance to complexity of the environment and evaluating the ability of the method in balancing between searching unexplored operating filed to safely move toward the target waypoint. For this purpose, a distinctive number of runs are performed to analyze the performance of the method in satisfying the problem constrains for all three mentioned scenarios.

The BBO optimization configuration for all scenarios set as follows: the habitats population (*nPop*) and maximum number of iterations (*Iter*$_{max}$) is set on 50 and 100, respectively. The number of kept habitats set on 10, number of new habitats is set on 40. The emigration rate is generated by μ=linspace(1,0,*nPop*), and the immigration rate defined as λ=1- μ. The maximum mutation rate is set on 0.1. Number of control points for each B-spline path is set on 8. Accuracy of the algorithm is tested for all scenarios and presented by *Fig.*8 to *Fig.*10 for different number of obstacles. The path should be re-generated simultaneously to avoid crossing the corresponding collision boundaries; and this process is repeated until vehicle reaches to the target waypoint.

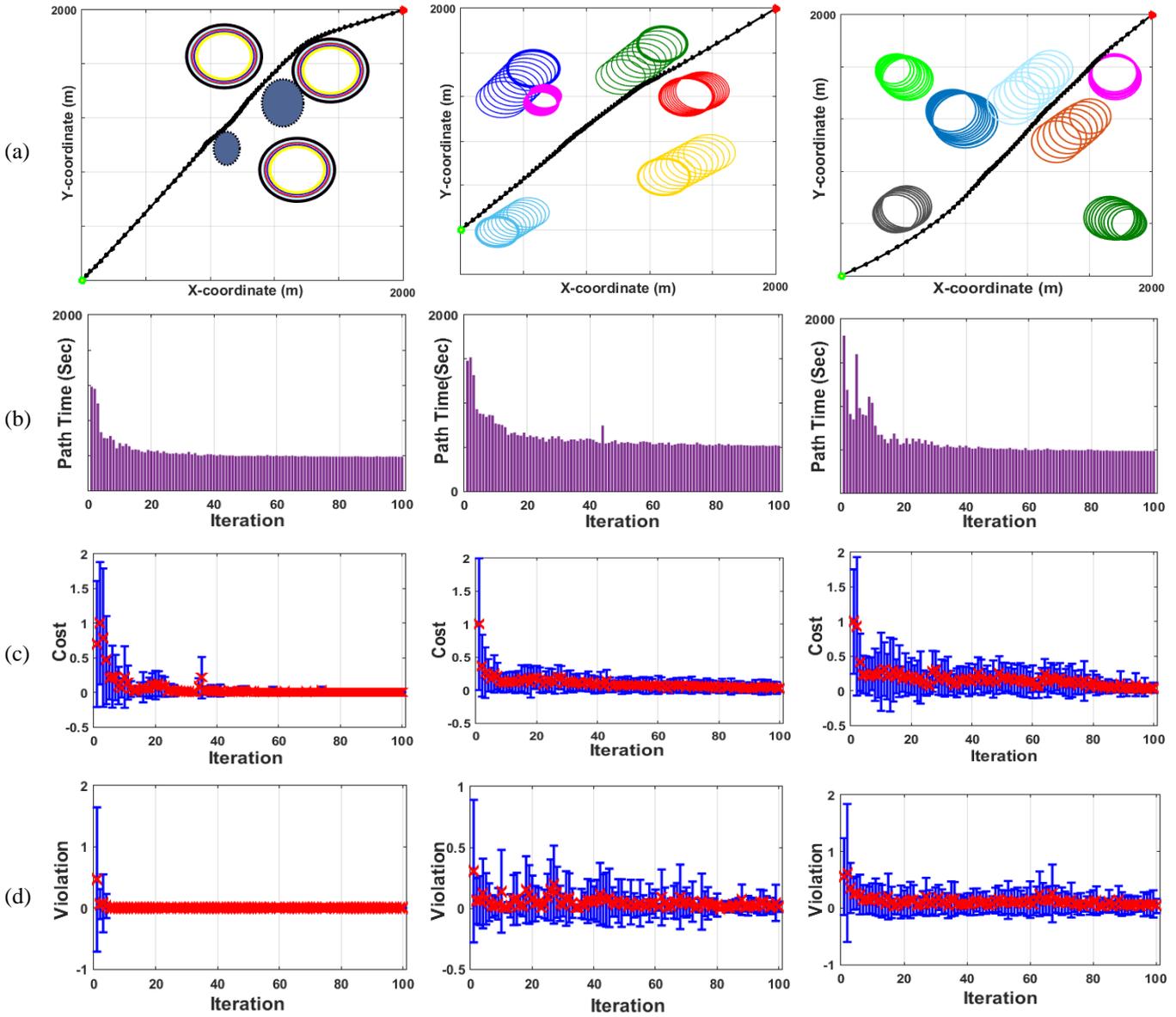

**Fig.8. (a)** 2D presentation of generated optimum 3D path including random combination of static known and static uncertain obstacles, **(b)** Path flight time variation in each iteration as an optimization factor, **(c)** Algorithms performance on cost variation of path population in each iteration, **(d)** Violation variation of path population in each iteration as collision penalty.

**Fig.9. (a)** 2D presentation of generated optimum 3D path including moving obstacles. **(b)** Path flight time variation in each iteration as an optimization factor, **(c)** Algorithms performance on cost variation of path population in each iteration, **(d)** Violation variation of path population in each iteration as collision penalty.

**Fig.10. (a)** 2D presentation of generated optimum 3D path including moving obstacles affected by current flow, **(b)** Path flight time variation in each iteration as an optimization factor, **(c)** Algorithms performance on cost variation of path population in each iteration, **(d)** Violation variation of path population in each iteration as collision penalty.

Figure.8 illustrates the performance of the path planner in the first scenario, where a random number of static obstacles with a growth in their radii are considered in the proposed operating field. The simulation results associated with the second scenario are demonstrated in *Fig*.9, where in *Fig*.9(a), the generated path in exposure of the moving obstacles is shown. The movement of obstacles is simulated in accordance with a pre-specified rate of uncertainty based on time. Considering the third scenario, the uncertainty around the obstacles in *Fig*.10(a) propagating from the centre of the object with a growth rate proportional to current velocity $|V_C|\sim N(0,0.3)$ in all directions. Additionally, the obstacles move with a self-motivated velocity in a random direction. The gradual increment of collision boundary of each obstacle has been presented in *Fig*.10(a). As can be seen from the *Fig*.10(a), the local path planner is able to generate a collision-free path even in a dynamic operating field with an acceptable rate of convergence for the defined optimality metric like flight time, shown in *Fig*.10(b). Albeit, the variations in terms of cost and violation functions per iteration are more significant when they get compared with the first and second scenarios, however, the algorithm still experiences a moderate convergence that guarantees feasible and optimal solutions.

It is further inferred from *Fig*.8(c,d), *Fig*.9(c,d), and *Fig*.10(c,d), the algorithm accurately tends to minimise path travel time and path cost over 100 iterations, while the performance of the algorithm is almost stable against increasing the complexity of the environment. It is also noteworthy to mention from analyze the results, the cost variation range in all scenarios decreases in each iteration which means algorithm accurately converges to the optimum solution with minimum cost. Tracking the variation of the mean cost and mean violation, that is represented by the red crosses in the middle of the error bar graphs, declares that algorithm enforces the solutions to approach the optimum answer (path) with minimum cost and efficiently manages the path to eliminate the collision penalty within 100 iterations.

In summary, it is obvious that in all three mentioned scenarios, the algorithm provides the solutions that satisfy the collision constraints for all types of obstacles; it tends to minimise the path travel time as it is the main optimisation factor for the path planner.

Additional to addressed common performance criterion investigated above, two more performance factors are highlighted for the purpose of this research that are discussed along the evaluation of the entire model. The first highlighted index is the path planner's computational time, because it must operate concurrently and synchronize to the global route planner. Hence, a large computational time causes the local path planner gets delayed in synchronisation with the graph route planner, which cause interruption in to the routine process of the whole system.

### B. Evaluation of the Global Route Planner

A number of performance metrics have been investigated to evaluate the optimality of the proposed solutions by the route planner in different network topologies, such as number of completed tasks, total obtained weight, total cost, and the time optimality of the generated route. Reliability percentage of the route is another metric representing chance of the mission success; it is defined based on route violation that is a weighted function of travel time and feasibility of the route. The BBO, in this circumstance, is configured with habitat population size of 150, iterations of 300, habitats keep rate of 0.6, emigration rate of $\mu=0.2$ immigration rate of $\lambda=1-\mu$, and maximum mutation rate is set on 0.8. Habitat keep rate is the ratio of the best solutions selected to be transferred to the next generation (as mentioned in the pseudo code in *Fig*.2). The performance of the algorithm is tested on graphs with diverse topologies including the graphs with 20, 50, 80 and 100 nodes. Table 1 demonstrates the impact of the graph complexity on functionality of the route planner.

TABLE 1. Statistical analyzing of the route (solution) evaluation with performance metrics for different graph complexities

| Performance metrics | Solution 1 | Solution 2 | Solution 3 | Solution 4 |
|---|---|---|---|---|
| **Graph Complexity-Nodes** | **20** | **50** | **80** | **100** |
| **Graph Complexity-Edges** | 202 | 1197 | 3099 | 4886 |
| **CPU Run Time** | 12.558 | 16.177 | 22.432 | 27.487 |
| **Best Cost** | 0.0413 | 0.0231 | 0.0196 | 0.0171 |
| **Total Mission Time(sec)** | 21200 | 32400 | 39600 | 45360 |
| **Route Travel Time(sec)** | 20848 | 30062 | 34686 | 45387 |
| **Total Distance** | 62543 | 90187 | 104059 | 136161 |
| **Total Weight** | 40 | 55 | 63 | 78 |
| **N-Tasks** | 15 | 19 | 23 | 27 |
| **Violation** | 0.0 | 0.0 | 0.0 | 0.008 |

Violation index gets a positive value when the route time exceeds the available time. Violating the cost function facilitates the algorithm to generate feasible solutions. It is noted from simulation results in Table 1, the violation value for all four examined complexities is almost zero, which means the generated routes accurately respect the defined time constraint. On the other hand, the optimum route corresponds to a route that has the closest travelling time to the total mission time, which means the AUV took maximum use of the total available time. As depicted in Table.1, in all four cases the route travel time is remarkably approached the value of the total mission time but not exceed it, which confirms the efficiency of the route planner in satisfying addressed objectives and constraints.

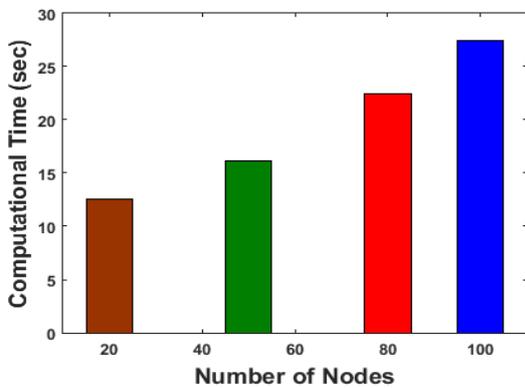
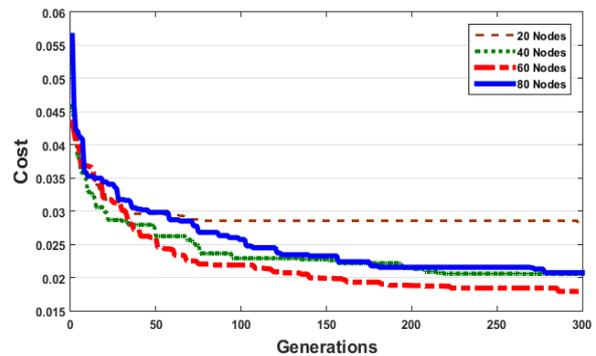

**Fig.11** Computational time variation to graph complexity.   **Fig.12** Cost variation of the BBO-based route planning for different graph complexities in 300 iterations.

Considering the influence of the graph topology on the presented solutions in *Fig*.11, the run time increases linearly as the number of nodes in graph increase, but in all cases remain within the bounds of a stable real-time solution. From the cost variations in *Fig.*12, it is obvious that the performance of the proposed route planner is stable to increasing size of the search space, which is again a major challenge for deterministic strategies that makes them inappropriate for real-time applications.

## C. Evaluation of the Combinatorial Local and Global Motion Planner Framework

The proposed combinatorial framework aims to take the maximum use of the mission available time to increase the number of completed tasks that hold higher priority in a single mission, while guarantying on-time termination of the mission and vehicles safe deployment during the mission. Accurate synchronization of the inputs and outputs of the engaged path/route planners and concurrent cooperation of them are the most important requirements in stability of the model toward the main objectives addressed above. To this end, the robustness of the model is evaluated through the simulation of 10 underwater missions in 10 individual experiments with the initial conditions that closely matches actual underwater mission scenarios. The initial configuration of the operation network has been set on 40 waypoints and 1320 edges involving a fixed sequence of tasks with specified characteristics (priority and completion time) in 10 $km^2$ (x-y),100 $m$ (z) space. The waypoint locations are initialized using uniform distribution of $\sim U(0,10000)$ for $WP^i_{x,y}$ and $\sim U(0,100)$ for $WP^i_z$. The mission available time for all experiments has been fixed on $T_{Available}$=10800(*sec*) equaling to 3 (*hours*). The vehicle starts its mission at initial location $WP^1$ and end its mission at $WP^{40}$. As mentioned earlier, it is assumed the vehicles is moving with constant (3 *m/s*) velocity. The operating field is modelled as a realistic underwater environment to examine the performance and stability of the proposed architecture. In Table.2 (A-B) the process of the combinatorial strategy in different stages of a specific mission scenario is shown.

Table. 2: An overview of the process of the combinatorial model in one mission scenario

**A. Global-Route**

| Call NO | Start | Target | Task NO | Weight | Route Cost | $T_{CPU}$ | $T_{Available}$ | $T_{Route}$ | Validity | Route Sequence |
|---|---|---|---|---|---|---|---|---|---|---|
| 1 | 1 | 40 | 8 | 38 | 0.430 | 23.1 | 10800 | 10460 | Yes | 1-24-7-25-32-11-26-34-40 |
| 2 | 25 | 40 | 6 | 27 | 0.320 | 20.9 | 5862.8 | 5831 | Yes | 25-36-26-27-33-5-40 |
| 3 | 33 | 40 | 1 | 14 | 0.610 | 19.8 | 1757.6 | 1728 | Yes | 33-40 |

**B. Local-Path**

| Route ID | PP Call NO | Edges | Violation | Path Cost | $T_{CPU}$ | $T_{path\text{-}flight}$ | $T_{Expected}$ | $T_{Available}$ | Replan Flag | PP Flag |
|---|---|---|---|---|---|---|---|---|---|---|
| *Route-1* | 1 | 1-24 | 0.000000 | 0.450 | 41.6 | 1532.7 | 1666.7 | 9267.3 | 0 | 1 |
|  | 2 | 24-7 | 0.000000 | 0.510 | 40.3 | 1702.3 | 1872.7 | 7565 | 0 | 1 |
|  | 3 | 7-25 | 0.000000 | 0.460 | 36.4 | 1702.1 | 1673.1 | 5862.8 | 1 | 0 |
| *Route-2* | 1 | 25-36 | 0.000031 | 0.115 | 37.8 | 467.2 | 535.3 | 5395.6 | 0 | 1 |
|  | 2 | 36-26 | 0.000000 | 0.311 | 43.7 | 1153.2 | 1210 | 4242.4 | 0 | 1 |
|  | 3 | 26-27 | 0.000000 | 0.369 | 39.1 | 1306.2 | 1333.4 | 2936.3 | 0 | 1 |
|  | 4 | 27-33 | 0.000007 | 0.232 | 39.7 | 1178.7 | 1068.3 | 1757.6 | 1 | 0 |
| *Route-3* | 1 | 33-40 | 0.000000 | 0.511 | 40.6 | 1705.4 | 1727.7 | **52.1** | 0 | 0 |

The mission starts with calling the global route planner for the first time; thus, a valid optimum route is generated to take maximum use of available time (valid route $T_{Route} \leq T_{Available}$). Referring to Table.1, the initial optimum route encapsulates number of 8 tasks with total weight of 38, cost of 0.430 and estimated completion time of $T_{Route}$=10460 (*sec*). In the second phase, the local path planner (PP) is recalled to generate optimum collision free path through the listed sequence of waypoint included in the initial route. Referring to Table.2(B), the local path planner adopts the first pair of waypoints {1-24} and generated optimum path between location of $WP^1$ and location of $WP^{24}$ with total cost of the 0.450 and travel time of $T_{path\text{-}flight}$=1532.7(*sec*) which is smaller than expected travel time $T_{Expected}$=1666.7(*sec*). The $T_{Expected}$ for the local path planner is calculated based on estimated travel time for the generated route ($T_{Route}$). In cases that $T_{path\text{-}flight}$ is smaller than $T_{Expected}$, the replanning flag is zero - the initial optimum route is still valid and optimum- so the vehicle is allowed to follow the next pair of waypoints included in initial optimum route. After each run of the path planner, the path time $T_{path\text{-}flight}$ is reduced from the total available time $T_{Available}$. The second pair of waypoints {24-7} is shifted to the path planner and the same process is repeated. However, if $T_{path\text{-}flight}$ exceeds the $T_{Expected}$, (occurred in {7-25}), the replanning flag gets one, which means some of the available time is wasted in passing the distance between $WP^7$ and $WP^{25}$ due to collision avoidance. In such a case also the $T_{Available}$ gets updated and visited edges (1-24, and 24-7) get eliminated from the graph. Afterward, instead of local path planner, the global route planner is recalled to generate new optimum route according to updated operation network and $T_{Available}$ from the current waypoint $WP^{25}$ to the predefined destination $WP^{40}$. In simulation results presented in Table.2, the global route planner is recalled for 3 times and the local path planner is called for 8 times within 3 optimum routes. This interplay among the modules continues until vehicle reaches to the destination (success) or $T_{Available}$ gets a minus value (failure: vehicle runs out of battery). In this case, the final route is the sequence of waypoints including {1-24-7-25-36-26-27-33-40} with total cost of 1.128, total weight of 31, and total time of 10748.8.

The most appropriate outcome for a mission is completion of the mission with the minimum remained time which means maximizing the use of mission available time. Referring Table.2 (B), the remaining time is 52.1 (*sec*) comparing to mission available time of $T_{Available}$=10800 (*sec*), which is remarkably approached to zero. Therefore, the architectures performance can be represented by mission time (or remained time). It is noteworthy that ensuring the on-time termination of the mission is prior for the proposed model than maximizing the completed tasks in a single mission, which is a big concern for vehicles safety; hence, a large penalty value is assigned to global route planner to strictly prevent generating routes with $T_{Route}$ bigger than $T_{Available}$. The time management performance of the model through 10 experiments is shown in *Fig.*13 and *Fig.*14.

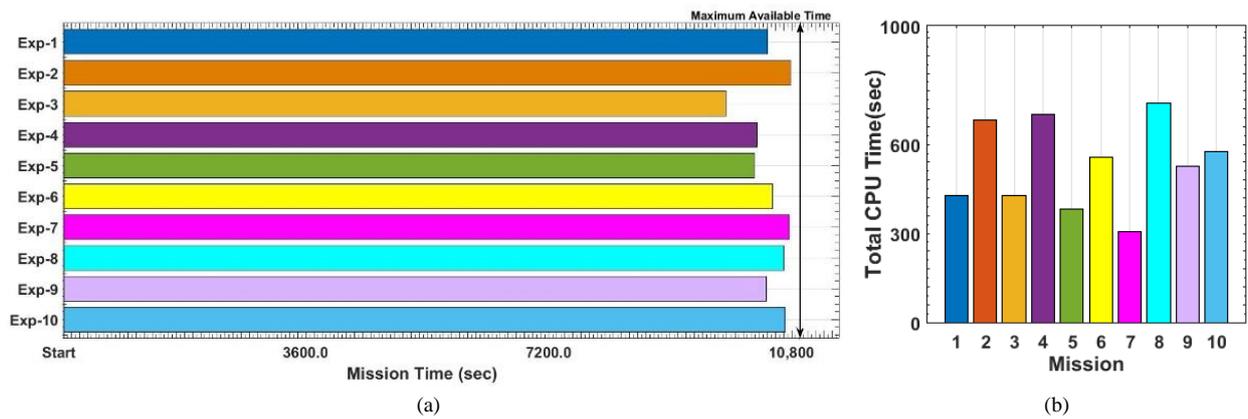

(a)                          (b)

**Fig.13 (a)** Performance of the combinatorial model in maximizing mission's performance by maximizing the mission time constrained to available time threshold, **(b)** Total operation's CPU time for 10 different experiments

As is apparent from *Fig.*13(a), the proposed model is capable of taking maximum use of mission available time as apparently the mission time in all experiments approaches the $T_{Available}$. It is important to mention that all generated solutions meet the constraints, denoted by an upper bound of 10800(*sec*); this declares the stability of the proposed strategy for any arbitrary mission in terms of time management and satisfying the mission objectives. Moreover, all missions are completed with a low computation burden as indicated by CPU time index in *Fig.*13(b). Besides, the model's cost variation is demonstrated in *Fig.*14 for each motion planner in all 10 experiments. It is obvious that the variation of the local path planner and global route planner average costs, obtained in each experiment, is centralized over the total cost of the final route generated in the corresponding mission that confirms stability of the models performance.

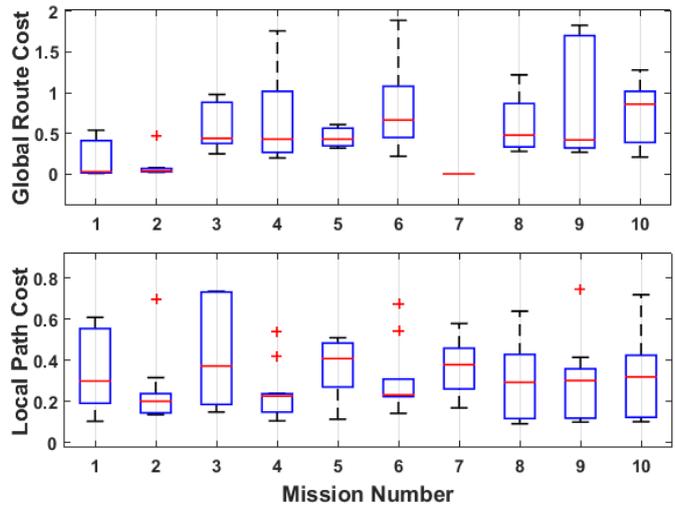

**Fig.14.** Model's stability on motion planners cost variation for 10 experiments

### 7 Conclusion

In this research a novel deliberative framework has been developed to raise the AUV potential for having a certain degree of autonomy in order to trade-off between tasks completion, time management, performing robust motion planning to successfully complete a mission. In the top level of the framework, the graph route planner tends to promote vehicles autonomy in terms of time management and task assignment in a graph-like terrain in which each edge of the graph is assigned with a task. Accordingly, in the lower level, the path planner tends to find shortest collision free trajectory between each pair of listed waypoints in the generated global route.

This approach can efficiently respond to environmental changes and is executable for real-time implementation. The stability and performance of the framework has been verified based on simulation results for various experiments. It is clear from the results that the presented model is efficient and accurate in producing real-time near optimal solution in which the efficiency of the model is relatively independent of both size and complexity of the operating field.

Future work will comprise more extensive version of the proposed architecture that is established upon the actual sensory information and estimation of one step ahead of environmental changes as it is useful in reality. Besides, the experimental validation of the framework on a real vehicle will be taken into account.